\documentclass[letterpaper, 10 pt, conference]{ieeeconf}
\IEEEoverridecommandlockouts
\usepackage{amsmath,amsthm,amssymb,amsfonts,mathtools}
\usepackage{graphicx}
\usepackage{textcomp}
\usepackage{caption}
\usepackage{subcaption} 
\usepackage{hyperref}
\hypersetup{
    colorlinks=true,
    linkcolor=blue,
    filecolor=magenta,      
    urlcolor=cyan,
}
\usepackage{xcolor}

\DeclareMathOperator*{\argmax}{argmax}

\DeclareMathOperator*{\sat}{sat}

\DeclareMathOperator*{\E}{\mathbb{E}}

\newcommand{\abs}[1]{\left|#1\right|}
\newtheorem{prop}{Proposition}
\newtheorem*{remark}{Remark}

\newcommand{\set}[1]{\left\{#1\right\}}
\newcommand{\brk}[1]{\left(#1\right)}
\newcommand{\bsq}[1]{\left[#1\right]}

\newcommand{\prob}[1]{\mathbb P\brk{#1}}
\newcommand{\R}{\mathbb{R}}

\DeclareMathAlphabet{\mymathbb}{U}{BOONDOX-ds}{m}{n}

\begin{document}

\title{Interaction-Aware Decision-Making for Autonomous Vehicles in Forced Merging Scenario Leveraging Social Psychology Factors\\
\thanks{$^{*}$Xiao Li, Kaiwen Liu, Anouck Girard, and Ilya Kolmanovsky are with the Department of Aerospace Engineering, University of Michigan, Ann Arbor, MI 48109, USA. {\tt\small \{hsiaoli, kwliu, anouck, ilya\}@umich.edu}}

\thanks{$^{\dagger}$H. Eric Tseng      Retired Senior Technical Leader, Ford Research and Advanced Engineering,             Chief Technologist, Excelled Tracer LLC.
{\tt\small hongtei.tseng@gmail.com}}

\thanks{This research was supported by the University of Michigan / Ford Motor Company Alliance program.}
}

\author{Xiao Li$^{*}$, Kaiwen Liu$^{*}$, H. Eric Tseng$^{\dagger}$, Anouck Girard$^{*}$, Ilya Kolmanovsky$^{*}$}

\maketitle

\begin{abstract}
Understanding the intention of vehicles in the surrounding traffic is crucial for an autonomous vehicle to successfully accomplish its driving tasks in complex traffic scenarios such as highway forced merging. In this paper, we consider a behavioral model that incorporates both social behaviors and personal objectives of the interacting drivers. Leveraging this model, we develop a receding-horizon control-based decision-making strategy, that estimates online the other drivers' intentions using Bayesian filtering and incorporates predictions of nearby vehicles' behaviors under uncertain intentions. The effectiveness of the proposed decision-making strategy is demonstrated and evaluated based on simulation studies in comparison with a game theoretic controller and a real-world traffic dataset.

\end{abstract}

\section{Introduction}\label{sec:intro}
One of the major challenges in autonomous driving lies in ensuring safe interaction with nearby traffic, particularly in highway merging scenarios. Unlike the simple stop-and-go strategy used in urban intersections with stop signs, the on-ramp ego vehicle must cooperate with high-speed vehicles and transition itself into the highway traffic in a secure, but also timely manner. Moreover, the varied social behaviors of different drivers can result in diverse responses to the merging intent. In this dynamic interaction, the ego vehicle must actively search for available space or create opportunities for merging. A cooperative driver on the highway may decelerate or change lanes to facilitate the merging process, while a self-interested driver may maintain a constant speed and disregard the merging vehicle. Consequently, understanding the driving intentions of the surrounding vehicles becomes crucial for the ego vehicle to accomplish its task successfully.

Learning-based methods have been extensively explored for autonomous driving applications. Without explicit interaction behavior modeling, Reinforcement-Learning (RL) based methods have been utilized to learn end-to-end driving policies~\cite{hugle2020dynamic, mavrogiannis2020b}. Additionally, researchers have employed imitation learning to train decision-making modules~\cite{pan2017agile, mei2021autonomous} that emulate expert behavior, such as a Model Predictive Controller~\cite{pan2017agile}. A comprehensive survey of RL methods in autonomous driving research can be found in~\cite{kiran2021deep}. However, a significant drawback of end-to-end RL-learned policies is the lack of interpretability; furthermore, their ability to generalize may be limited
by the interactive behaviors observed in the training data.


To address this challenge, researchers have explored the integration of learning-based methods with planning and control techniques. The Inverse Reinforcement Learning (IRL) approach has been employed to learn the reward function of human drivers for planning purposes~\cite{sadigh2016planning, you2019advanced}. Additionally, neural network models, such as the Social Generative Adversarial Network~\cite{gupta2018social}, have been implemented in trajectory prediction modules for Model Predictive Control (MPC)~\cite{espinoza2022deep,bae2022lane}. Novel network architectures have also been designed to enhance driving motion forecasting~\cite{liang2020learning,ivanovic2020multimodal}. However, a common issue in these learned modules is their limited generalization capability beyond the training dataset.

Game theoretic approaches have also been considered to represent interactions between agents in traffic, such as the Level-k method~\cite{li2016hierarchical}, potential games~\cite{liu2022potential}, and Stackleberg games~\cite{hang2020human, fisac2019hierarchical}. A Leader-Follower Game theoretic Controller (LFGC) has been proposed specifically for modeling pairwise leader-follower interactions in forced merging scenarios in~\cite{liu2022interaction, liu2021cooperation}. Inspired by the concepts presented in~\cite{liu2022interaction, liu2021cooperation}, we adopt a pairwise interaction formulation to model vehicle cooperative behaviors, enabling better scalability for scenarios involving multiple vehicles.

Differently from~\cite{liu2022interaction, liu2021cooperation}, to create a more comprehensive model of human driving, we incorporate a novel behavioral model that incorporates various social psychology factors. Early studies in social psychology have revealed that individuals don't always act solely to maximize their own rewards in two-person~\cite{kuhlman1975individual} or $n$-person experimental games~\cite{liebrand1984effect}. Drawing inspiration from the concept of Social Value Orientation (SVO)~\cite{liebrand1984effect, messick1968motivational, liebrand1986might, griesinger1973toward}, and its application in the context of autonomous driving~\cite{DanielaRusSVO}, we propose a novel behavioral model that encompasses both the drivers' inclination towards social cooperation and their individual objectives as latent parameters. Leveraging this proposed behavioral model, we can estimate the underlying driving intentions of the interacting vehicles and make appropriate decisions for the ego vehicle in forced merging scenarios. The algorithms we propose offer several potential advantages:
\begin{enumerate}
    \item The proposed behavioral model incorporates aspects of both driver's social cooperativeness and personal objectives and captures a rich and realistic set of behaviors.
    \item The algorithm uses a Bayesian filter to infer the latent driving intent parameters, thereby handling uncertainties in the cooperation intent of interacting vehicles.
    \item The derived decision-making module adopts a pairwise interaction formulation and utilizes receding-horizon optimization-based control that leads to good scalability while ensuring safety for forced merging applications. 
\end{enumerate}

This paper is organized as follows: In Sec.~\ref{sec:problem}, we introduce the problem setting and the forced merging scenario. We also outline the assumptions made regarding vehicle kinematics, action space, and driver's action objectives. In Sec.~\ref{sec:methodModel}, we present our behavioral model that incorporates the cooperation intents and personal objectives of the interacting vehicles. In Sec.~\ref{sec:methodCtrl}, we present the decision-making module for the ego vehicle, that effectively handles uncertainties in the driving intentions of the interacting vehicles. In Sec.~\ref{sec:results}, we demonstrate the ability of our behavioral model to reproduce realistic driving behaviors. Furthermore, we validate the proposed controller through simulations in comparison with the LFGC and real-world dataset evaluations. Finally, the conclusions are given in Sec.~\ref{sec:conclusion}.
\section{Problem Formulation}\label{sec:problem}
\begin{figure}[!h]
    \centering
    \includegraphics[width=0.47\textwidth]{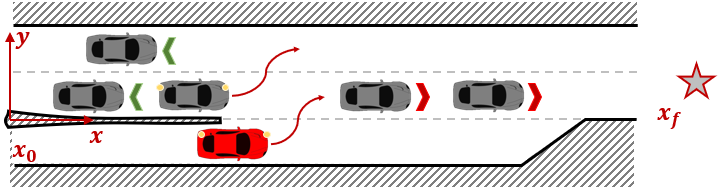}
    \caption{Schematic diagram of the highway forced merging problem: an ego vehicle (red) interacts with highway vehicles (grey) to facilitate its merging.}
    \label{fig:forcedMergingProblem}
\end{figure}

In this paper, we focus on the design of a decision-making module for autonomous driving applications in forced merging scenarios. The decision-making module plans high-level behaviors such as acceleration, deceleration, or lane changing, and generates desired reference trajectories for the autonomous vehicle. Subsequently, a lower-level controller is assumed to be available that can control the steering and acceleration/braking of the vehicle to track the reference trajectory. As illustrated in Fig.~\ref{fig:forcedMergingProblem}, the goal is to design a behavior planner for the ego vehicle to merge into the target highway lane while accounting for interactions with multiple highway vehicles to ensure safe and effective merging. 

\subsection{Vehicle Kinematics Model}\label{subsec:vehicleKine}
We use the following discrete-time model to represent the vehicle kinematics,
\begin{equation}\label{eq:EoM}
    \left[\begin{array}{c}
        x(t+1)\\
        v_x(t+1)\\
        y(t+1)
    \end{array}\right]
    = \left[\begin{array}{c}
        x(t) + v_x(t)\Delta t\\
        v_x(t) + a(t)\Delta t\\
        y(t) + v_y(t)\Delta t\\    
    \end{array}\right] + \tilde{w}(t),
\end{equation}
where $x$, $v_x$, and $a$ are the longitudinal position, velocity, and acceleration, respectively; $y$ and $v_y$ are the lateral position and velocity; $\Delta t > 0$ is the sampling period between discrete time instances $t$ and $t+1$; $\tilde{w}(t)\in\R^3$ is a disturbance representing unmodeled dynamics.

We assume all the vehicles, including both ego and highway vehicles, follow this dynamics model. For simplicity, Eq.~\eqref{eq:EoM} can be rewritten as,
\begin{equation}\label{eq:fEoM}
    s_i(t+1) = f\big(s_i(t), u_i(t)\big)  + \tilde{w}_i(t),\; i = 0, 1, 2,\dots,
\end{equation}
where $s_i(t)=[x(t), y(t), v_x(t) ]^T$ and $u_i(t)=[a(t), v_y(t)]^T$ represent the state and control of the $i$'th vehicle at time instance $t$, respectively. In the following context, the subscript $i=0$ designates the ego vehicle, while $i\in\set{1,2,\dots}$ represents another vehicle with which the ego vehicle interacts.

\subsection{Action Space}\label{subsec:actionSpace}
We assume vehicles take actions from the action set $U$ that comprises:
\begin{enumerate}
    \item ``Maintain": keep the current lateral position and longitudinal speed;
    \item ``Accelerate": keep the current lateral position and accelerate at $a\;\rm m/s^2$ without exceeding the upper speed limit $v_{\text{max}}\;\rm m/s$;
    \item ``Decelerate": keep the current lateral position and decelerate at $-a\;\rm m/s^2$ without falling below the lower speed limit $v_{\text{min}}\;\rm m/s$;
    \item ``Steer to the left": keep the current longitudinal speed and steer to the left adjacent lane with a constant lateral velocity of $\frac{w_{\text{lane}}}{T_{\text{lane}}}\;\rm m/s$;
    \item ``Steer to the right": keep the current longitudinal speed and steer to the right adjacent lane with a constant lateral velocity of $-\frac{w_{\text{lane}}}{T_{\text{lane}}}\;\rm m/s$.
\end{enumerate}
Note that we assume a complete lane change takes $T_{\text{lane}}$ sec to move into the adjacent lane with a lateral traveling distance of lane width $w_{\text{lane}}$, which is reflected by the above actions. We also note that more acceleration and deceleration levels $\abs{a}$ can be introduced to the action space, but we only consider one level here for simplicity.

\subsection{Driving Objectives}\label{subsec:objectives}
The driving objectives of each vehicle are reflected in the reward function that depends on the following four variables:
\begin{enumerate}
    \item Traffic rules $c$: The binary variable $c\in\set{0,1}$ is an indicator for either getting into a collision with other vehicles or getting beyond the road boundaries. A safety bounding box is constructed that overbounds each vehicle body in the $x-y$ plane with certain safety margins. The value of $c = 1$ indicates the overlap of two vehicles' bounding boxes or the overlap between the vehicle's bounding box and the road boundary. The value of $c=0$ indicates that the vehicle stays within the road and is not in collision with other vehicles. The visualization of the highway road boundaries is shown in Fig.~\ref{fig:forcedMergingProblem} using solid black lines.
    \item Safety consciousness $h$: The variable $h\in[0,1]$ is derived from the Time-to-Collision $(TTC)$ with a vehicle ahead in the same lane
    \begin{equation*}
        h = \frac{\sat_{\bsq{T_{\text{min}}, T_{\text{max}}}} (TTC)- T_{\text{min}}}{T_{\text{max}} - T_{\text{min}}},
    \end{equation*}
    where $T_{\text{min}}=0.2$ sec is the minimum reaction time, $T_{\text{max}}=3$ sec stands for an adequate time headway, and $\sat_{\bsq{a,b}}\brk{\cdot}$ is a saturation function between the minima $a$ and the maxima $b$. The reward function depends on $h$ to encourage vehicles to keep an appropriate headway distance and be conscious of potential collisions. 
    \item Traveling time $\tau$: The variable $\tau \in [0,1]$ reflects the objective of shortening the traveling time, and is a weighted summation of 
    \begin{equation*}
        \tau_x = \frac{x - x_0}{x_f - x_0}, \tau_y =1 - \frac{1}{w_{\text{lane}}}\min\brk{\abs{y-y_r}, w_{\text{lane}}},
    \end{equation*}
    where $x_0$ and $x_f$ (see Fig.~\ref{fig:forcedMergingProblem}) are the $x-$coordinates of the beginning of the ramp and a goal placed a specified distance away from the end of the ramp, respectively, while $y_r$ corresponds to the center of the highway lane that is next to the ramp. The reward for $\tau_x$ promotes the highway vehicle reaching the end of the highway in a shorter time. A higher reward for $\tau_y$ is imposed for on-ramp vehicles to encourage merging action.
    \item Control effort $e$: The reward for $e \in [0, 1]$ promotes vehicles to drive at a constant speed and to reduce acceleration/deceleration. The variable $e$ attains value of 1 under the action ``maintain"; its value decreases if the vehicle makes speed changes or lane changes. 
    
\end{enumerate}

\section{Social Behavior Modeling}\label{sec:methodModel}
In this section, we introduce our behavioral model that captures drivers' interactive decision-making process during the forced merge scenario. Inspired by social psychology studies~\cite{liebrand1984effect, messick1968motivational, liebrand1986might, griesinger1973toward}, and its application in the context of autonomous driving~\cite{DanielaRusSVO}, we define the SVO-based reward model in Sec.~\ref{subsec:reward}. In Sec.~\ref{subsec:behaviorModel}, we integrate this reward model into the interacting vehicle’s decision-making process.


\subsection{Social Value Orientation and Multi-modal Reward}\label{subsec:reward}
We assume each vehicle $i$ interacts pairwise with each adjacent vehicle $j\in A(i)$, where $A(i)$ contains indices of all the adjacent vehicles around $i$. We assume each driver aims to achieve their personal objectives and, to a certain extent, is cooperating with others. Hence, we model each driver's intention using a multi-modal reward function of the form
\begin{equation}\label{eq:reward_R}
\begin{array}{l}
    R_i \big(s(t), u(t)| \sigma_i, w_i \big) = \frac{1}{\abs{A(i)}} \sum_{j\in A(i)} \cdot
    \\ \Big[ \theta_1(\sigma_i) \cdot  r_i \big(s_i(t),u_i(t),s_j(t),u_j(t) | w_i \big) 
    \\+ \theta_2(\sigma_i)\cdot r_j \big(s_j(t),u_j(t),s_i(t),u_i(t) | w_j \big) \Big],
\end{array}
\end{equation}
where $u(t)=[u_i^T(t), u_{A(i)}^T(t)]^T$ is the aggregated control vector of all vehicles and $u_{A(i)}(t)$ is a column vector concatenating $u_j(t)$ for all $j\in A(i)$; $s(t)=[s_i^T(t), s_{A(i)}^T(t)]^T$ reflects the state of the traffic at time $t$; $\abs{A(i)}$ is the number of interacting vehicles; $r_i(\cdot)$ and weights $w_i\in\R^3$ model personal reward as a weighted summation of personal objectives defined in Sec.~\ref{subsec:objectives},
\begin{equation}\label{eq:reward_r}
   r_i(s_i,u_i,s_j,u_j | w_i)  = (\neg c)\cdot w_i^T \cdot [h,\tau, e]^T.
\end{equation}
The symbol $\neg$ in Eq.~\eqref{eq:reward_r} is the logical negative operator and $\sigma_i$ in Eq.~\eqref{eq:reward_R} takes one of four values corresponding to four SVO categories and specific values of $\theta_1(\sigma_i)$ and $\theta_2(\sigma_i)$:
\begin{equation}\label{eq:sigma}
    (\theta_1, \theta_2) =\left\{
    \begin{array}{cl}
         (0,1) &  \text{if } \sigma_i = \text{ altruistic}\\
         (1/2, 1/2) &  \text{if } \sigma_i = \text{ prosocial}\\
         (1,0) &  \text{if } \sigma_i = \text{ egoistic}\\
         (1/2, -1/2) &  \text{if } \sigma_i = \text{ competitive}\\ 
    \end{array}
    \right..
\end{equation}
Note that in Eq.~\eqref{eq:reward_R}, $\theta_1$ and $\theta_2$ correspond to the weight of the self-reward $r_i$ and the weight of the other drivers' net reward, respectively. 

In this multi-modal reward given by Eq.~\eqref{eq:reward_R}, there are two latent parameters $(\sigma_i, w_i)$ that represent different driving incentives: $w_i$ reflects different personal goals, and $\sigma_i$ represents different social behaviors or levels of cooperativeness. For instance, a driver with weights $w_i = [0,1,0]^T$ in Eq.~\eqref{eq:reward_r} might consider driving at full speed thereby minimizing the traveling time. As implied by Eq.~\eqref{eq:reward_R}, a ``prosocial" driver has equal weights between personal objectives and other drivers' objectives; hence such drivers intend to cooperate with others in pursuing a large net reward. Note that $w_j$ is the internal parameter of vehicle $j$ and is a latent variable affecting the decision of vehicle $i$ if $\sigma_i\neq \text{``egoistic"}$ and $j\in A(i)$. Nonetheless, an altruistic or prosocial (or competitive) driver of vehicle $i$ is likely to improve (or diminish) other drivers' rewards in all three variables if they do not know other drivers' objectives $w_j$ a priori. Therefore, we assume that during the $i$'th vehicle's decision-making, $w_j=[1/3, 1/3, 1/3]$ in Eq.~\eqref{eq:reward_R} for $j \in A(i)$.

\subsection{Driving Behavior Model}\label{subsec:behaviorModel}
In our behavior model, we assume the driver of vehicle $i$ aims to maximize the cumulative reward, defined as
\begin{equation}\label{eq:behavior_model_Q_prime}
\begin{array}{l}     
    Q'_i \big(s(t), \gamma_i|\sigma_i,w_i \big) = \\
    \E_{\gamma_j, j\in A(i)}\bsq{ \sum\limits_{k=0}^{N-1} \lambda^k R_i \big( s(t+k), u(t+k) \big| \sigma_i,w_i \big) },
\end{array}
\end{equation}
where $\gamma_i=\set{u_i(t+k)}_{k=0}^{N-1}\in U^N$ is an action sequence over a horizon of length $N$, and $\lambda\in[0,1]$ is a discount factor. This cumulative reward is an averaged reward over all possible action sequences $\gamma_j$ of vehicles $j\in A(i)$. Furthermore, the driver of vehicle $i$ is assumed to adopt a receding horizon control strategy, i.e., 
\begin{equation}\label{eq:behavior_model_MPC}
    u^*_i(t) = \argmax\limits_{u\in U} Q_i \big(s(t), u|\sigma_i,w_i \big),
\end{equation}
where $Q_i (s(t), u)=\mathbb{E}_{\gamma_i\in\Gamma^1(u)}\bsq{Q'_i(s(t), \gamma_i|\sigma_i,w_i)}$ and $\Gamma^1(u)=\set{\gamma_i=\set{u_i(t+k)}_{k=0}^{N-1}: u_i(t)=u}$ contains all the action sequences with the initial action $u$. Furthermore, considering stochasticity in the decision-making process, a policy distribution can be prescribed by adopting a softmax decision rule \cite{sutton2018reinforcement}:
\begin{equation}\label{eq:behavior_model_policy}
    \mathbb{P} \big(u_i = u | \sigma_i, w_i, s(t)\big) \propto \exp \big( Q_i (s(t), u|\sigma_i,w_i) \big).
\end{equation}

Based on the behavioral model defined above, the model parameters $\sigma_i,w_i$ can affect action policies to represent different driving intentions. Since other drivers’ intentions are not known in a given traffic scenario, the model parameters $\sigma_i, w_i$ (i.e., drivers' intentions) need to be estimated and updated online so that the autonomous ego vehicle is able to make optimal merging decisions.
\section{Decision-Making Under Cooperation Intent Uncertainty}\label{sec:methodCtrl}
We now develop a decision-making algorithm to facilitate the forced merging process. We first present a Bayesian filter for the ego vehicle that estimates the latent variables $\sigma_i,w_i$ of the interacting vehicles online. Considering the uncertainties in our estimation and other drivers' intentions, we use a receding-horizon control formulation to simultaneously address the safety and performance aspects of the forced merging.

\subsection{Bayesian Inference of Latent Driving Intentions}\label{subsec:inference}

At each time step, we assume that the ego vehicle can observe the traffic nearby the $i$'th interacting vehicle where $i\in A(0)$, and the observed traffic history is defined as 
\begin{equation*}
\begin{array}{l}    
    \xi(t) =\big\{s(0), s(1), \dots, s(t), \\
    u_{A(i)}(0), u_{A(i)}(1), \dots, u_{A(i)}(t-1)\big\},
\end{array}
\end{equation*}
where $s(t)=[s_i^T(t), s_{A(i)}^T(t)]^T$ and $u_{A(i)}(t)$ is a column vector concatenating $u_j(t)$ for all $j\in A(i)$. The ego vehicle utilizes the traffic history to estimate the posterior distribution $\prob{\sigma_i,w_i| \xi(t+1)}$ of the $i'$th interacting vehicle's latent parameters using the following proposition:

\begin{prop}
Given a prior $\prob{\sigma_i,w_i|\xi(t)}$ and assuming that the disturbance $\tilde{w}_i(t)\sim\mathcal{N}(0, Q)$ is zero-mean Gaussian, the posterior distribution can be computed as 
\begin{equation}\label{eq:Bayesian}
    \prob{\sigma_i,w_i| \xi(t+1)} = \frac{\Lambda_i\brk{\sigma_i,w_i, s(t), s_i}}{N_i(t)}\cdot\prob{\sigma_i,w_i| \xi(t)},
\end{equation}
where $N_i(t)$ is a normalization factor and $\Lambda_i\brk{\sigma_i,w_i, s(t), s_i}$ admits the following form:
\begin{equation}\label{eq:transiton_prob}
\begin{array}{l}     
    \Lambda_i\brk{\sigma_i,w_i, s(t), s_i} = \sum\limits_{u\in U}\; 
    \prob{u_i = u | \sigma_i, w_i, s(t)} \cdot
    \\ \prob{\tilde{w}_i(t) = s_i - f(s_i(t), u)},
\end{array}
\end{equation}
where $\prob{u_i = u | \sigma_i, w_i, s(t)}$ is defined in Eq.~\eqref{eq:behavior_model_policy}. 
\end{prop}

Note that the above recursive Bayesian filter can be initialized using a uniform distribution and the covariance matrix $Q$ is a tunable parameter. Intuitively, if we consider the current traffic state $s(t)$ and the vehicle $i$ is executing policy defined in Eq.~\eqref{eq:behavior_model_policy} conditioned on parameters $\sigma_i$ and $w_i$, $\Lambda_i\brk{\sigma_i,w_i, s(t), s_i}$ represents the transition probability of the vehicle $i$ moving from $s_i(t)$ to $s_i(t+1)=s_i$. The proof is presented as follows: 

\begin{proof}
Applying the Bayesian rule, the posterior admits the following form,
\begin{equation*}
    \begin{array}{l}
    \prob{\sigma_i,w_i| \xi(t+1)} = \prob{\sigma_i,w_i| s(t+1), u_{A(i)}(t), \xi(t)}
    \\=\frac{\prob{s(t+1)| \sigma_i,w_i, u_{A(i)}(t), \xi(t)}}{\prob{s(t+1)| u_{A(i)}(t), \xi(t)}}\cdot\prob{\sigma_i,w_i| \xi(t)}.
\end{array}
\end{equation*}
Moreover, the likelihood term can be rewritten as follows,
\begin{equation*}
    \begin{array}{l}
    \prob{s(t+1)| \sigma_i,w_i, u_{A(i)}(t), \xi(t)}
    \\=\prob{s_{A(i)}(t+1)| u_{A(i)}(t), s_{A(i)}(t)}\cdot
    \\ \prob{s_i(t+1)| \sigma_i,w_i,s(t)}
\end{array},
\end{equation*}
whereby the posterior reduces to 
\begin{equation*}
\begin{array}{l}
    \prob{\sigma_i,w_i| \xi(t+1)} \propto\\
    \prob{s_i(t+1)| \sigma_i,w_i,s(t)}\cdot \prob{\sigma_i,w_i| \xi(t)},
\end{array}
\end{equation*}
and $\Lambda_i\brk{\sigma_i,w_i, s(t), s_i}=\prob{s_i(t+1)| \sigma_i,w_i,s(t)}$ is the transition probability conditioned on the model parameters $\sigma_i$, $w_i$ and the current traffic state $s(t)$.
\end{proof}

\subsection{Receding-horizon Optimization-based Control}\label{subsec:control}

We leverage the receding horizon control to achieve safe merging. The objective of successful merging without collisions can be translated into maximizing predictive cumulative reward,
\begin{equation}\label{eq:ego_Q_prime}
\begin{array}{l}
    Q'_0(s(t), \gamma_0) 
    = \frac{1}{\abs{A(0)}}\sum\limits_{i\in A(0)}  \E\limits_{\sigma_i,w_i\sim \prob{\sigma_i,w_i| \xi(t)}} \\
    \set{
    \E\limits_{\gamma_i \sim \prob{\gamma_i | \sigma_i, w_i, s(t)}}
     \bsq{  \overline{Q}'_0 \big(s_0(t), s_i(t), \gamma_0, \gamma_i \big) }
    },\\
\end{array}
\end{equation}
where $A(0)$ contains indices of all the vehicles near the ego vehicle; $\gamma_0$, $\gamma_i$ are action sequences of length $N$ corresponding to the ego vehicle and the vehicle $i$; $\prob{\sigma_i,w_i| \xi(t)}$ is the latent parameter estimated distribution and is updated by the Bayesian filter; Similar to Eq.~\eqref{eq:behavior_model_policy}, $\prob{\gamma_i | \sigma_i, w_i, s(t)}$ is an augmented policy distribution derived from Eq.~\eqref{eq:behavior_model_Q_prime} such that $\prob{\gamma_i| \sigma_i, w_i, s(t)}\propto \exp\brk{Q'_i(s(t), \gamma_i|\sigma_i,w_i)}$; $\overline{Q}'_0\brk{\cdot}$ computes the discounted cumulative reward of the ego vehicle according to
\begin{equation*}
\begin{array}{l}
    \overline{Q}'_0(s_0(t), s_i(t), \gamma_0, \gamma_i)\\    
    = \sum\limits_{k=0}^{N-1} \lambda^k r_0(s_0(t+k),u_0(t+k),s_i(t+k),u_i(t+k) ),
\end{array}
\end{equation*}
and the one-step reward function $r_0(s_0,u_0,s_i,u_i)  = (\neg c)\cdot \tau$ encourages collision avoidance and faster merging. Note that we can also customize the reward function $r_0$ for other implementation requirements, e.g., add variable $\tau$ of other drivers to $r_0$ to mitigate traffic congestion introduced by the ego vehicle's merging action.

Here, we model the ego vehicle interactions pairwise with the nearby vehicles $i\in A(0)$. For each interaction vehicle $i$, the outer expectation in Eq.~\eqref{eq:ego_Q_prime} considers all possible combinations of parameters $\sigma_i,w_i$ that illustrate the cooperation intent and personal objectives of the interacting vehicle $i$. The inner expectation then takes an average return of all plausible decisions by vehicle $i$ conditioned on the parameters $\sigma_i,w_i$ under our behavior model. The reward function $\overline{Q}'_0\brk{\cdot}$ represents the cumulative gain of the pairwise interaction between the ego vehicle and the vehicle $i$. 

Based on the cumulative reward, the ego vehicle is controlled using a receding horizon control strategy, i.e., 
\begin{equation}\label{eq:ego_MPC}
    u^*_0(t) = \argmax\limits_{u\in U} Q_0 (s(t), u),
\end{equation}
where $Q_0 (s(t), u)=\mathbb{E}_{\gamma_0\in\Gamma^1(u)}\bsq{Q'_0(s(t), \gamma_0)}$ and $\Gamma^1(u)$ contains all action sequences of length $N$ with $u$ being their initial action. 
\begin{remark}
    Note that the reward computation in Eq.~\eqref{eq:ego_Q_prime} is not reactive. Our algorithm does not predict other vehicles' reactions to the ego actions in the prediction. To address this concern, we can also adopt a game-theoretic formulation similar to \cite{liu2022interaction, liu2021cooperation, DanielaRusSVO}. However, such a formulation is computationally demanding in practice. Due to the formulation of pairwise interaction in Eq.~\eqref{eq:ego_Q_prime}, our algorithm can solve Eq.~\eqref{eq:ego_MPC} effectively via an exhaustive search that scales linearly with the number of interacting vehicles.
\end{remark}

\section{Simulation and Experimental Results}\label{sec:results}
Here, we demonstrate the effectiveness of the proposed behavioral model and the forced merging control algorithm. The behavioral model is first demonstrated by reproducing real-world driving behaviors. Then, the effectiveness of the proposed forced merging control algorithm is illustrated through simulation studies in comparison with the LFGC~\cite{liu2022interaction, liu2021cooperation} against interacting vehicles controlled by our behavior model and through a naturalistic driving dataset present in High-D \cite{highD}.

\begin{figure*}[ht!]
    \centering
    \vspace{0.5em}
    \includegraphics[width=0.9\textwidth]{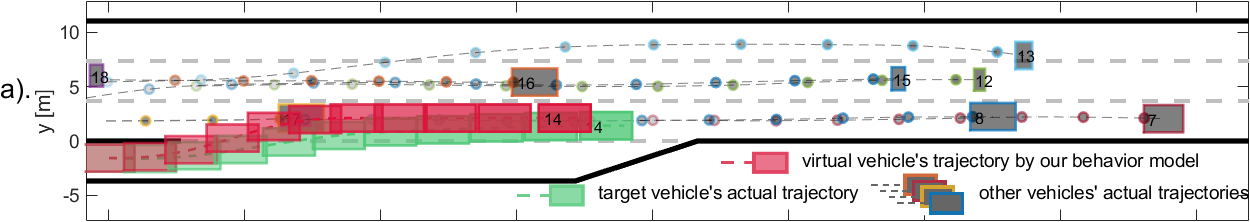}
    \includegraphics[width=0.9\textwidth]{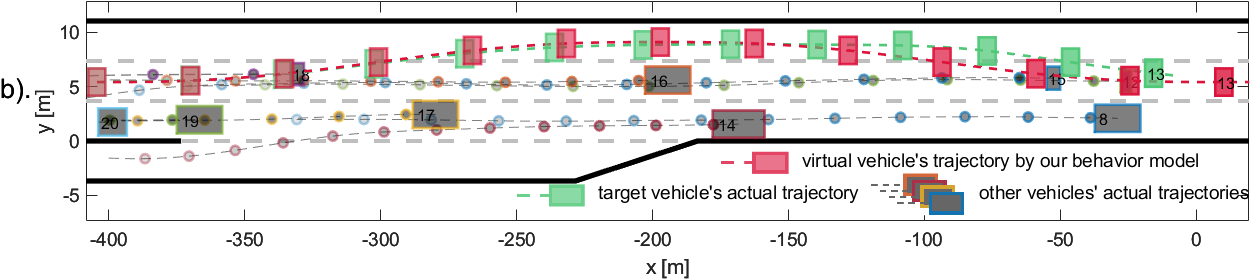}
    \caption{\textbf{Examples of reproducing real-world highway merging and overtaking behaviors}: In each example, the target vehicle is represented by a green box, and interacts with the traffic. All traffic vehicles are visualized as boxes of different edge colors filled with grey. The trajectory of vehicles is shown in dashed lines with vehicles' positions every second marked as boxes in green (for the target vehicle) and circles filled with grey (for other traffic vehicles). The virtual vehicles' trajectories are visualized using red dashed lines and red boxes and match the actual ones closely. a) A traffic of 12 seconds is sampled from frames 1-300 in scene 59 from the High-D dataset. The trajectory of vehicle 14 is reproduced using our behavioral model with $\sigma_{14}=\text{egoistic},$ $w_{14}=[0, 2/3, 1/3]^T$, i.e., vehicle 14 is an egoistic driver, and minimizes the traveling time by merging to the highway. b) A traffic of 14 seconds is sampled from frames 1-350 in scene 59 from the High-D dataset. The trajectory of vehicle 13 is reproduced using our behavioral model with $\sigma_{13}=\text{competitive},$ $w_{13}=[0, 1, 0]^T$, i.e., vehicle 13 is a competitive driver, and minimizes the traveling time by overtaking the leading vehicles.}
    \label{fig:reproducingExample}
\end{figure*}

\subsection{Reproducing Real-world Traffic}\label{subsec:result_reproducing}
\begin{figure}[ht!]
    \centering
    \includegraphics[width=0.47\textwidth]{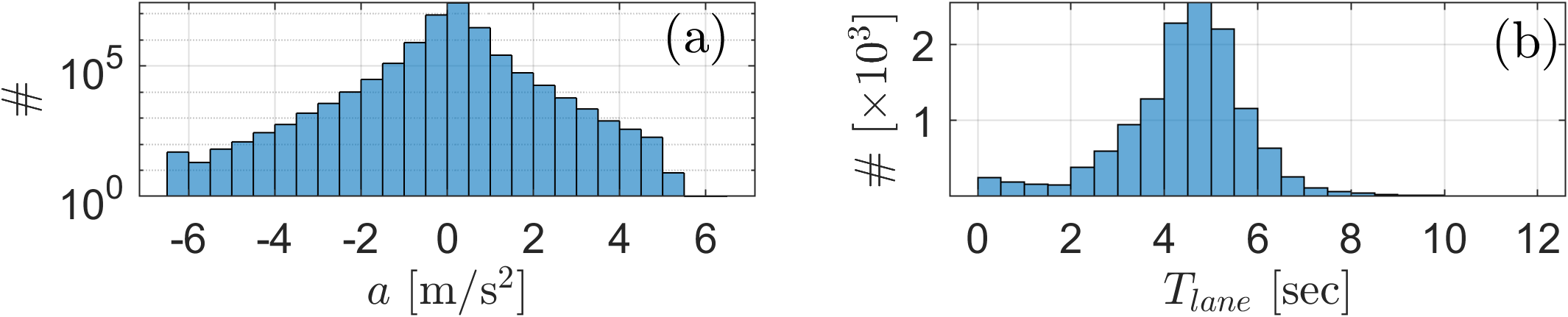}  
    \caption{Histogram of vehicle driving statistics in High-D dataset: (a) Longitudinal acceleration/deceleration (y-axis in log scale). (b) Time duration for a complete lane change.}
    \label{fig:param_histogram}
\end{figure}

We leverage a drone-recorded naturalistic traffic dataset called the High-D dataset \cite{highD}. It contains 60 traffic recordings among which three recordings/scenes (58-60) have merging ramps. We first calibrate the values of the model parameters $a$ and $T_{\text{lane}}$ from the High-D dataset (see Fig.~\ref{fig:param_histogram}). The lane width $w_{\text{lane}}=3.5\;\rm m$ in High-D. We choose $\abs{a} = 6\;\rm m/s^2$ because the longitudinal accelerations and decelerations of High-D vehicles are within an interval of $[-6,6]\;\rm m/s^2$. And we select $T_{\text{lane}}=4\;\rm sec$ since the majority of the vehicles in the High-D dataset take $4\sim 6\;\rm sec$ to change lane.

We aim to reproduce the real-world driving behavior (see Fig.~\ref{fig:reproducingExample}) in the recording segments using our behavioral model with certain parameters $\sigma_i, w_i$. Specifically, to reproduce the behavior of a target vehicle $i$, we initialize our behavioral model with its actual initial state $s_i(t=0)$ from the recording segment. At each time step of 2 seconds, we assume a virtual vehicle $i$ is controlled by our behavior model, and can observe the surrounding traffic $s(t) = [s_i(t)^T, s_{A(i)}(t)^T]^T$. With prescribed parameters $\sigma_i, w_i$, the virtual vehicle $i$ updates its control decision every time step using the behavioral model defined in Eq.~\eqref{eq:behavior_model_MPC} with MPC prediction horizon $N=3$ spanning 6 seconds. The kinematics of the virtual vehicle obeys Eq.~\eqref{eq:EoM} with a sampling rate of 25 $\rm Hz$. Furthermore, to introduce realistic lateral behaviors, the underlining lane change trajectories are modeled by 5th-order polynomials as proposed in~\cite{TomiPolyLaneChange}.

As shown in Fig.~\ref{fig:reproducingExample}a), the virtual truck $14$ in green is controlled by our behavioral model with parameters $\sigma_{14}=\text{egoistic},$ $w_{14}=[0, 2/3, 1/3]^T$. This combination of parameters implies the virtual vehicle cares about the self-reward solely by the formulation in Eq.~\eqref{eq:reward_R} and Eq.~\eqref{eq:sigma} and tries to minimize the traveling time because it has the largest weight $2/3$ for $\tau$ in Eq.~\eqref{eq:reward_r}. As a result, the virtual truck merges into the highway, and the trajectory in the green boxes matches the actual target vehicle in the red boxes. As shown in Fig.~\ref{fig:reproducingExample}b), in front of the virtual vehicle $13$, there are two vehicles $12$ and $15$ driving slowly. With parameters $\sigma_{13}=\text{competitive},$ $w_{14}=[0, 1, 0]^T$, the behavioral model controls the virtual vehicle to compete with the two vehicles while minimizing its traveling time. This interaction results in an overtaking behavior for the virtual vehicle which qualitatively matches the actual traffic recording. These examples provide evidence of our behavioral model being able to capture realistic driving behaviors. Note that there are quantitative mismatches in positions between the reproduced and the actual results in Fig.~\ref{fig:reproducingExample}. Such mismatches partially result from the model assumption in Eq.~\eqref{eq:EoM}; improvements to the kinematic model are left as future work.

\subsection{Forced Merging in Simulation Compared with LFGC}\label{subsec:result_sim}
\begin{figure}[h!]
    \centering
    \includegraphics[width=0.48\textwidth]{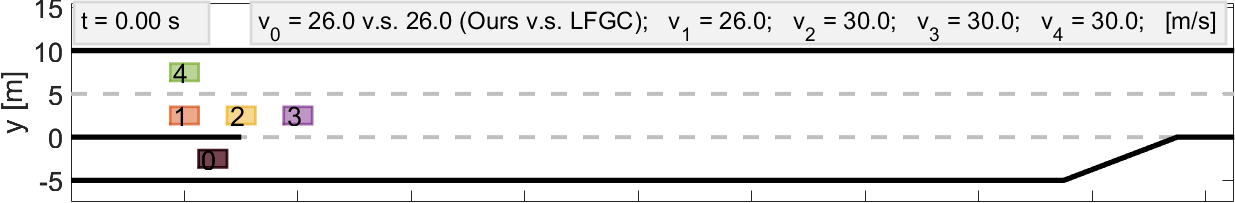}
    \includegraphics[width=0.48\textwidth]{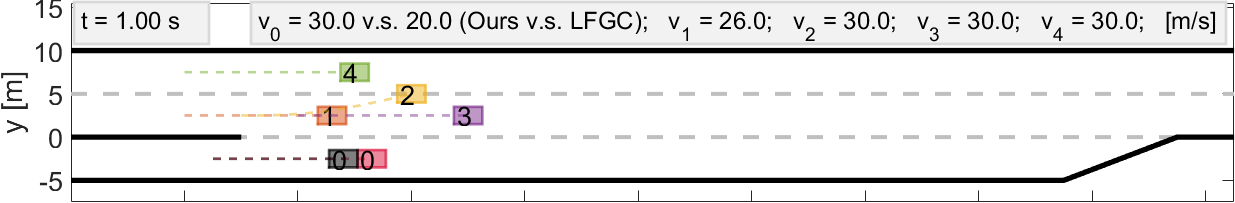}
    \includegraphics[width=0.48\textwidth]{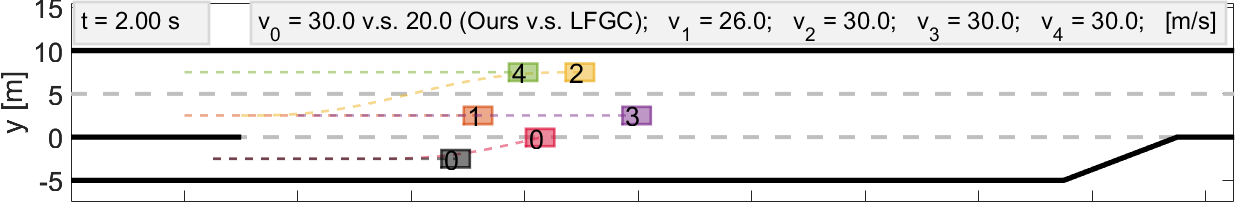}
    \includegraphics[width=0.48\textwidth]{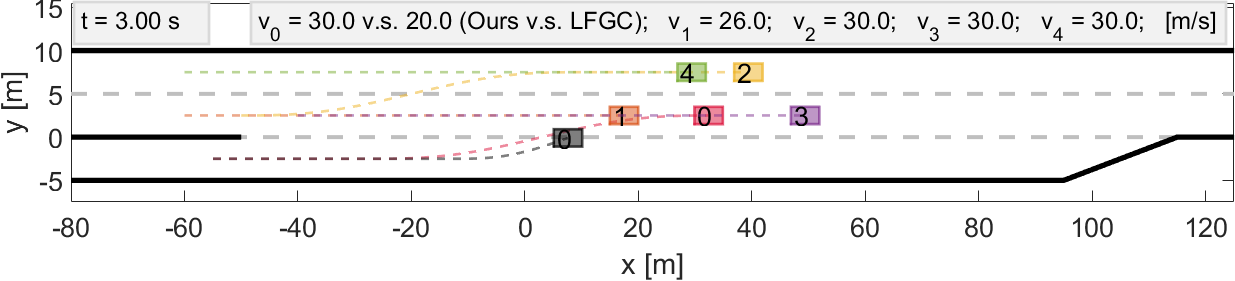} 
    \includegraphics[width=0.4\textwidth]{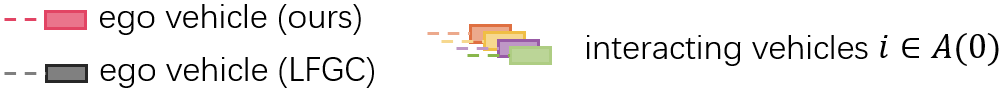}  
    \caption{Forced merging comparison in a simulation environment: Ego vehicle controlled by our algorithm in the red box is interacting with the adjacent vehicles in colored boxes. The LFGC is tested using the same highway traffic setting, and the results are plotted overlaid using grey boxes. Four sub-figures demonstrate the interactions at $t=0,1,2,3$ seconds, respectively. After accelerating to surpass vehicle 1, our ego vehicle properly merges into the highway at $t=3$ seconds.}
    \label{fig:compExampleTraj}
\end{figure}

We build up a highway with five vehicles (see Fig.~\ref{fig:compExampleTraj}) to simulate a forced merging scenario. The ego vehicle interacts with the surrounding traffic using the proposed controller in Eq.~\eqref{eq:ego_MPC}. In this example, each time step corresponds to one second and the lane change takes two time steps. Other vehicles are controlled using our behavioral model with different parameters $\sigma_i,w_i$. All the traffic vehicles are modeled as ``egoistic" drivers, i.e., $\sigma_i = $ ``egoistic" for all $i=1,\dots,4$. We assign vehicle $i=1,3,4$ with weights $w_i=[0,0,1]$ to encourage minimization of control effort $e$ such that the three vehicles will keep a constant speed. Vehicle 2 with weights $w_2=[1,0,0]$ tries to search for a larger headway space $h$, therefore, changes to the inner lane from $t=0$ to $t=2$ seconds.

As shown in Fig.~\ref{fig:compExampleTraj}, we also deploy the LFGC to control the ego vehicle in the same highway traffic setting, and the results are plotted overlaid with ours in comparison. Note that the parameters $\sigma_i$ and $w_i$ of all vehicles 1-4 are not available to ego vehicle 0. As a result, the ego vehicle needs to interact with other vehicles and estimate their intentions online to facilitate its merging. During the simulation, our ego vehicle successfully infers the cooperative intents of vehicle $1,2,3$ from their behaviors, and it decides to first accelerate to surpass vehicle $1$ from $t=0$ to $t=1$, and merges into the gap created by the lane changing of vehicle $2$. 

\begin{figure}[h!]
    \centering
    \vspace{0.5em}
    \includegraphics[width=0.48\textwidth]{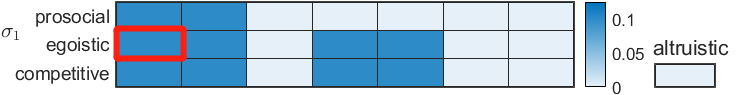}
    \includegraphics[width=0.48\textwidth]{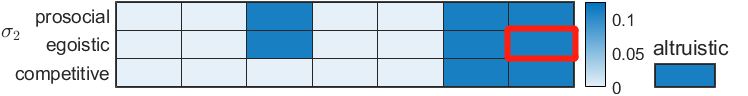}
    \includegraphics[width=0.48\textwidth]{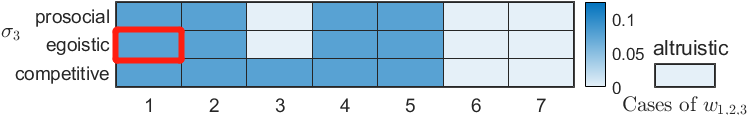}    
    \caption{Bayesian filter estimations of posterior $\prob{\sigma_i,w_i| \xi(t)}$: Three sub-figures visualize the parameter belief of the interacting vehicles $i=1,\;2,\;3$, respectively, with $t=1$ second. The $x$ axis labels 7 different weights $w_i\in W$ with order listed in Eq.~\eqref{eq:W_set}. The $y$ axis labels the SVO categories $\sigma_i$, with the ``altruistic" one standing alone to the right. The heatmap presents a higher probability with a deeper blue and the actual model parameters are circled in red.}
    \label{fig:compExampleFilter}
\end{figure}

We also provide our Bayesian filter estimation results (see Fig.~\ref{fig:compExampleFilter}) at time step $t=1$. Here, we confine the domain of the weights $w_i$ into a finite subset $W\subset[0,1]^3$ as follows
\begin{equation}\label{eq:W_set}
W = \set{
\begin{aligned}    
     [0,0,1],\; & [0,1,1]/2,\;   [0,1,0],\; [1,1,1]/3,\\
     [1,0,1]/2,\; & [1,1,0]/2,\; [1,0,0]
\end{aligned}
},
\end{equation}
where each weight case is a normalized combination of zeros and ones. Moreover, as shown in Fig.~\ref{fig:compExampleFilter}, the $\sigma_i=$``altruistic" category stands alone and is not correlated with the reward weights $w_i$ because the ``altruistic" driver does not care about the self-reward as modeled in Eqs.~\eqref{eq:reward_R},~\eqref{eq:reward_r} and~\eqref{eq:sigma}. Notably, the actual parameters $\sigma_i,w_i$ are among the ones of the highest probability. Meanwhile, for vehicle $2$, the cases with the highest probability mostly emphasize traveling time $\tau$ and headway distance $h$ while those of vehicles $1,3$ emphasize control effort $e$. Using these probability distributions, the ego vehicle can predict the driving and cooperation intent of vehicles $1,2,3$, and plan its task accordingly, as demonstrated in Fig.~\ref{fig:compExampleTraj}.

In comparison, the ego vehicle controlled by the LFGC estimates the probability of vehicles being a leader or being a follower, i.e., 
$$\prob{i=\text{``leader"}} = 1 - \prob{i=\text{``follower"}}, i=1,2,3,4.$$
However, due to the lane-changing behavior, the LFGC cannot distinguish between vehicle $2$ being a ``leader" and it being a ``follower", namely, $\prob{i=\text{``leader"}} = \prob{i=\text{``follower"}} = 0.5$. This results in a deceleration decision of the LFGC from $t=0$ to $t=1$. Seeing the constant-speed vehicle $1$ as a ``leader", the LFGC further keeps a low speed from $t=1$ to $t=2$ and decides to merge after vehicle $1$ at $t=2$ while the ego vehicle controlled by our algorithm is in the middle of a lane change. This provides evidence that, compared to the LFGC, our behavioral model captures a richer and more realistic set of behaviors and the controller integrated with our behavioral model can achieve faster merging in certain cases.

\subsection{Validation on Real-world Dataset}\label{subsec:result_highD}
\begin{figure}[h!]
    \centering
    \vspace{0.5em}
    \includegraphics[width=0.48\textwidth]{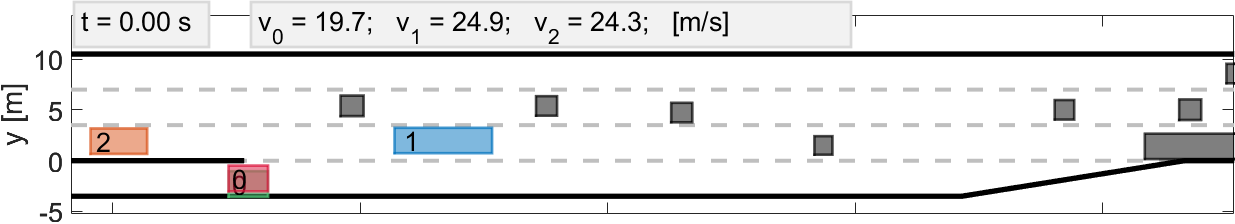}
    \includegraphics[width=0.48\textwidth]{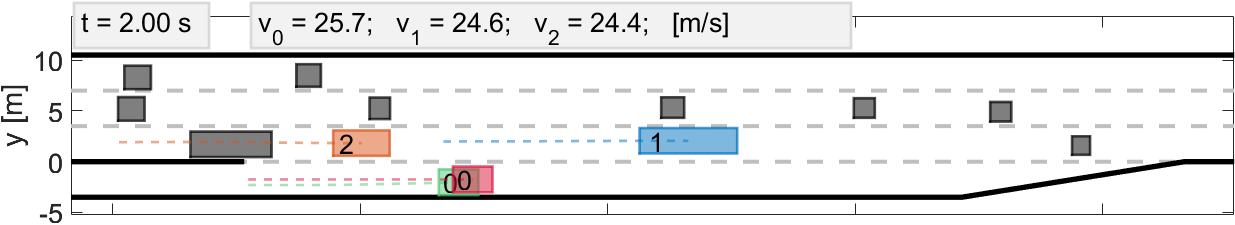}
    \includegraphics[width=0.48\textwidth]{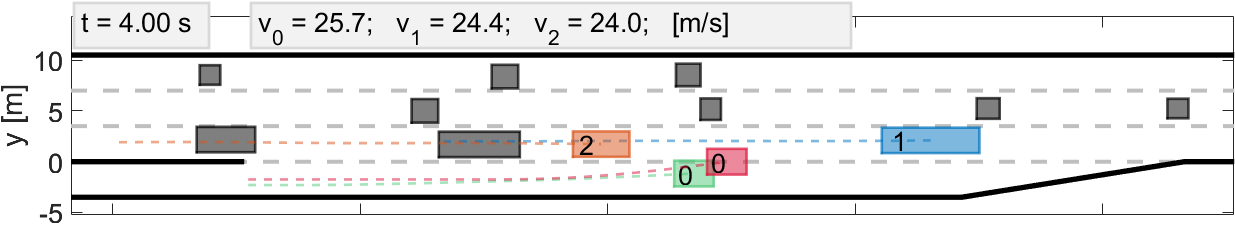}
    \includegraphics[width=0.48\textwidth]{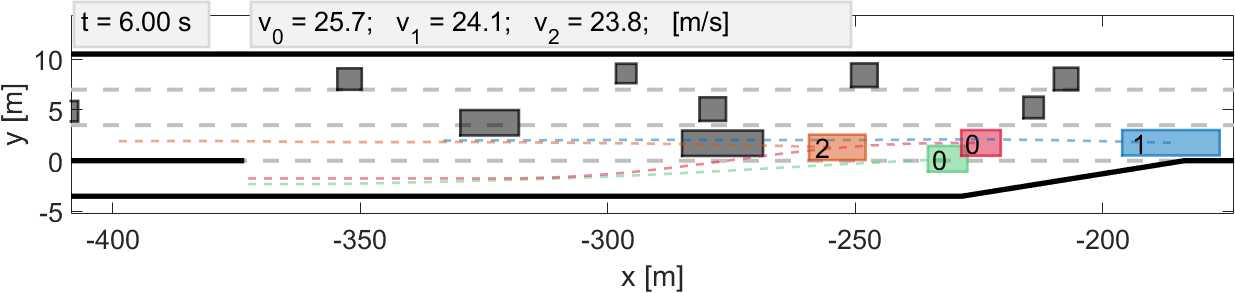} 
    \includegraphics[width=0.48\textwidth]{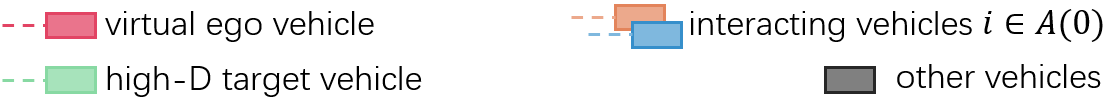}  
    \caption{An example of forced merging validation on the High-D dataset: the virtual ego vehicle in the red box is initialized using the target vehicle's initial state. Its adjacent vehicles are in colored boxes while other traffic vehicles are visualized in grey ones. Four sub-figures demonstrate the interactions at $t=0,2,4,6$ seconds, respectively. After accelerating, the virtual ego vehicle properly merges onto the highway.}
    \label{fig:highDExampleTraj}
\end{figure}

To validate our controller in real-world traffic, we consider traffic segments (see Fig.~\ref{fig:highDExampleTraj}) that contain merging vehicles from the High-D dataset. Similar to Sec.~\ref{subsec:result_reproducing}, we use $w_{\text{lane}}=3.5\;\rm m$, $\abs{a} = 6\;\rm m/s^2$, $T_{\text{lane}}=4\;\rm sec$, and an MPC prediction horizon $N=3$ spanning 6 seconds. We initialize our virtual ego vehicle using the initial state of the merging vehicle on the ramp. Afterward, we control the virtual ego vehicle with the proposed controller. We use the finite weight set $W$ in Eq.~\eqref{eq:W_set} for parameter estimation in the Bayesian filter. 

\begin{figure}[h!]
    \centering
    \includegraphics[width=0.48\textwidth]{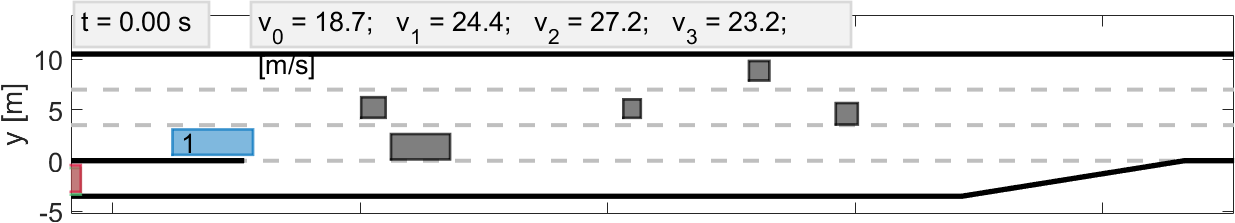}
    \includegraphics[width=0.48\textwidth]{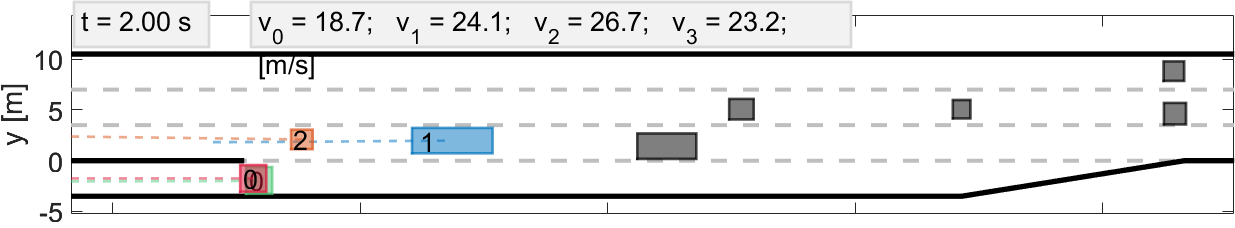}
    \includegraphics[width=0.48\textwidth]{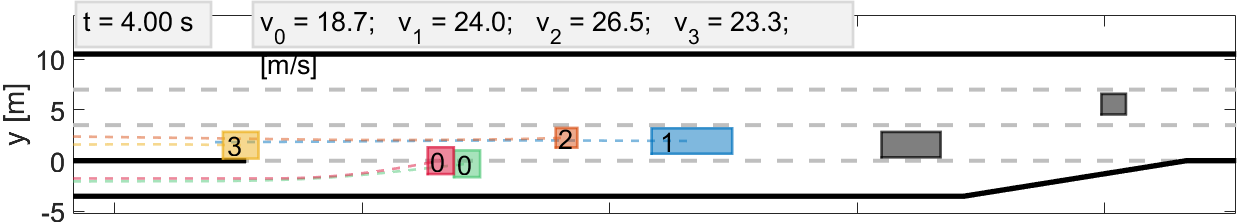}
    \includegraphics[width=0.48\textwidth]{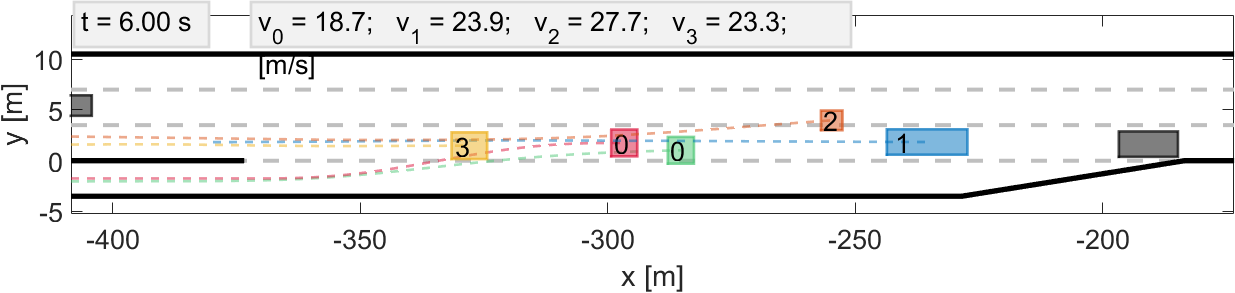} 
    \includegraphics[width=0.48\textwidth]{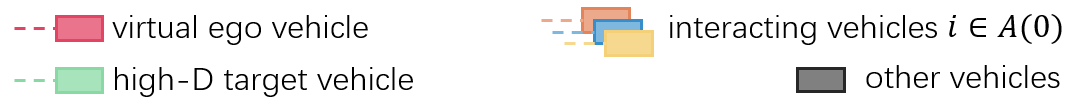}  
    \caption{Another example of forced merging validation on the High-D dataset: the virtual ego vehicle first keeps a constant speed, then merges onto the highway.}
    \label{fig:highDExampleTraj2}
\end{figure}

As shown in Fig.~\ref{fig:highDExampleTraj}, the virtual ego vehicle interacts with two trucks $1$ and $2$ that drive approximately at constant speeds. Thus, the ego vehicle first accelerates to create adequate merge space and speed advantage, then successfully merges between the two trucks. Another example is provided in Fig.~\ref{fig:highDExampleTraj2}, where the ego vehicle keeps a constant speed for 2 seconds to create merge space after vehicle 2. Notably, in both examples, we can observe that the ego vehicle's trajectories are similar to the ones of the actual target vehicles, shown in green boxes. 

Meanwhile, in the High-D dataset, there are in total of 75 merging vehicles in scenes 58-60. For each merging vehicle, we repeat the aforementioned procedures to set up the test environment and control the virtual ego vehicle to merge onto the highway. The test results are presented in Table~\ref{tab:highDResults}. We consider a test case a success if the ego vehicle successfully merges without collisions. A failure case implies that the ego vehicle either collides with other vehicles or fails to merge by the end of the ramp. Our algorithm can achieve a $100\%$ success rate among the $75$ test cases, and properly merges the virtual vehicles into the naturalistic traffic. 

\begin{table}[h!]
  \centering
    \vspace{0.5em}
  \caption{High-D forced merging test results on scenes 58, 59, and 60: the traffic recordings in three scenes are recorded during different time intervals on Wednesday 07/2018 in the same highway location.}
    \begin{tabular}{c|c|c|c|c}
    \hline\hline
    Scene Number & 58 & 59 & 60 & Total\\\hline
    Time Interval (AM) & 9:15-9:22 & 9:23-9:31 & 9:37-9:53 &  \\
    Number of Merges & 18 & 21 & 36 & 75\\
    Success & 18 & 21 & 36 & 75\\
    Failure & 0 & 0 & 0 & 0\\\hline
    Success Rate (\%) & 100 & 100 & 100 & 100\\
    \hline\hline    
    \end{tabular}%
  \label{tab:highDResults}
\end{table}
\section{Conclusion}\label{sec:conclusion}
In this paper, we proposed a driving behavioral model that takes different social value orientations and personal objectives into consideration. Based on the proposed behavioral model, we developed a Bayesian filter that estimates online the latent cooperation intent of the interacting vehicles, and proposed a control algorithm that simultaneously achieves the merging objective and ensures driving safety. Finally, we demonstrate the effectiveness of the proposed behavioral model and the forced merge control algorithm by reproducing real-world trajectories and evaluating the merging performance in simulations in comparison with the LFGC and a naturalistic traffic dataset.


\bibliographystyle{IEEEtran}
\bibliography{ref.bib}

\begin{thebibliography}{10}
\providecommand{\url}[1]{#1}
\csname url@samestyle\endcsname
\providecommand{\newblock}{\relax}
\providecommand{\bibinfo}[2]{#2}
\providecommand{\BIBentrySTDinterwordspacing}{\spaceskip=0pt\relax}
\providecommand{\BIBentryALTinterwordstretchfactor}{4}
\providecommand{\BIBentryALTinterwordspacing}{\spaceskip=\fontdimen2\font plus
\BIBentryALTinterwordstretchfactor\fontdimen3\font minus
  \fontdimen4\font\relax}
\providecommand{\BIBforeignlanguage}[2]{{%
\expandafter\ifx\csname l@#1\endcsname\relax
\typeout{** WARNING: IEEEtran.bst: No hyphenation pattern has been}%
\typeout{** loaded for the language `#1'. Using the pattern for}%
\typeout{** the default language instead.}%
\else
\language=\csname l@#1\endcsname
\fi
#2}}
\providecommand{\BIBdecl}{\relax}
\BIBdecl

\bibitem{hugle2020dynamic}
M.~H{\"u}gle, G.~Kalweit, M.~Werling, and J.~Boedecker, ``Dynamic
  interaction-aware scene understanding for reinforcement learning in
  autonomous driving,'' in \emph{International Conference on Robotics and
  Automation (ICRA)}.\hskip 1em plus 0.5em minus 0.4em\relax IEEE, 2020, pp.
  4329--4335.

\bibitem{mavrogiannis2020b}
A.~Mavrogiannis, R.~Chandra, and D.~Manocha, ``B-gap: Behavior-rich simulation
  and navigation for autonomous driving,'' \emph{Robotics and Automation
  Letters}, vol.~7, no.~2, pp. 4718--4725, 2022.

\bibitem{pan2017agile}
Y.~Pan, C.-A. Cheng, K.~Saigol, K.~Lee, X.~Yan, E.~Theodorou, and B.~Boots,
  ``Agile autonomous driving using end-to-end deep imitation learning,''
  \emph{arXiv preprint arXiv:1709.07174}, 2017.

\bibitem{mei2021autonomous}
X.~Mei, Y.~Sun, Y.~Chen, C.~Liu, and M.~Liu, ``Autonomous navigation through
  intersections with graph convolutionalnetworks and conditional imitation
  learning for self-driving cars,'' \emph{arXiv preprint arXiv:2102.00675},
  2021.

\bibitem{kiran2021deep}
B.~R. Kiran, I.~Sobh, V.~Talpaert, P.~Mannion, A.~A. Al~Sallab, S.~Yogamani,
  and P.~P{\'e}rez, ``Deep reinforcement learning for autonomous driving: A
  survey,'' \emph{Transactions on Intelligent Transportation Systems}, vol.~23,
  no.~6, pp. 4909--4926, 2021.

\bibitem{sadigh2016planning}
D.~Sadigh, S.~Sastry, S.~A. Seshia, and A.~D. Dragan, ``Planning for autonomous
  cars that leverage effects on human actions.'' in \emph{Robotics: Science and
  systems}, vol.~2.\hskip 1em plus 0.5em minus 0.4em\relax Ann Arbor, MI, USA,
  2016, pp. 1--9.

\bibitem{you2019advanced}
C.~You, J.~Lu, D.~Filev, and P.~Tsiotras, ``Advanced planning for autonomous
  vehicles using reinforcement learning and deep inverse reinforcement
  learning,'' \emph{Robotics and Autonomous Systems}, vol. 114, pp. 1--18,
  2019.

\bibitem{gupta2018social}
A.~Gupta, J.~Johnson, L.~Fei-Fei, S.~Savarese, and A.~Alahi, ``Social gan:
  Socially acceptable trajectories with generative adversarial networks,'' in
  \emph{Conference on Computer Vision and Pattern Recognition (CVPR)}, 2018,
  pp. 2255--2264.

\bibitem{espinoza2022deep}
J.~L.~V. Espinoza, A.~Liniger, W.~Schwarting, D.~Rus, and L.~Van~Gool, ``Deep
  interactive motion prediction and planning: Playing games with motion
  prediction models,'' in \emph{Learning for Dynamics and Control Conference
  (L4DC)}.\hskip 1em plus 0.5em minus 0.4em\relax PMLR, 2022, pp. 1006--1019.

\bibitem{bae2022lane}
S.~Bae, D.~Isele, A.~Nakhaei, P.~Xu, A.~M. A{\~n}on, C.~Choi, K.~Fujimura, and
  S.~Moura, ``Lane-change in dense traffic with model predictive control and
  neural networks,'' \emph{Transactions on Control Systems Technology}, 2022.

\bibitem{liang2020learning}
M.~Liang, B.~Yang, R.~Hu, Y.~Chen, R.~Liao, S.~Feng, and R.~Urtasun, ``Learning
  lane graph representations for motion forecasting,'' in \emph{European
  Conference on Computer Vision (ECCV)}.\hskip 1em plus 0.5em minus 0.4em\relax
  Springer, 2020, pp. 541--556.

\bibitem{ivanovic2020multimodal}
B.~Ivanovic, K.~Leung, E.~Schmerling, and M.~Pavone, ``Multimodal deep
  generative models for trajectory prediction: A conditional variational
  autoencoder approach,'' \emph{Robotics and Automation Letters}, vol.~6,
  no.~2, pp. 295--302, 2020.

\bibitem{li2016hierarchical}
N.~Li, D.~Oyler, M.~Zhang, Y.~Yildiz, A.~Girard, and I.~Kolmanovsky,
  ``Hierarchical reasoning game theory based approach for evaluation and
  testing of autonomous vehicle control systems,'' in \emph{Conference on
  Decision and Control (CDC)}.\hskip 1em plus 0.5em minus 0.4em\relax IEEE,
  2016, pp. 727--733.

\bibitem{liu2022potential}
M.~Liu, I.~Kolmanovsky, H.~E. Tseng, S.~Huang, D.~Filev, and A.~Girard,
  ``Potential game-based decision-making for autonomous driving,''
  \emph{Transactions on Intelligent Transportation Systems}, pp. 1--14, 2023.

\bibitem{hang2020human}
P.~Hang, C.~Lv, Y.~Xing, C.~Huang, and Z.~Hu, ``Human-like decision making for
  autonomous driving: A noncooperative game theoretic approach,''
  \emph{Transactions on Intelligent Transportation Systems}, vol.~22, no.~4,
  pp. 2076--2087, 2020.

\bibitem{fisac2019hierarchical}
J.~F. Fisac, E.~Bronstein, E.~Stefansson, D.~Sadigh, S.~S. Sastry, and A.~D.
  Dragan, ``Hierarchical game-theoretic planning for autonomous vehicles,'' in
  \emph{International Conference on Robotics and Automation (ICRA)}.\hskip 1em
  plus 0.5em minus 0.4em\relax IEEE, 2019, pp. 9590--9596.

\bibitem{liu2022interaction}
K.~Liu, N.~Li, H.~E. Tseng, I.~Kolmanovsky, and A.~Girard, ``Interaction-aware
  trajectory prediction and planning for autonomous vehicles in forced merge
  scenarios,'' \emph{Transactions on Intelligent Transportation Systems}, 2022.

\bibitem{liu2021cooperation}
K.~Liu, N.~Li, H.~E. Tseng, I.~Kolmanovsky, A.~Girard, and D.~Filev,
  ``Cooperation-aware decision making for autonomous vehicles in merge
  scenarios,'' in \emph{Conference on Decision and Control (CDC)}.\hskip 1em
  plus 0.5em minus 0.4em\relax IEEE, 2021, pp. 5006--5012.

\bibitem{kuhlman1975individual}
D.~M. Kuhlman and A.~F. Marshello, ``Individual differences in game motivation
  as moderators of preprogrammed strategy effects in prisoner's dilemma.''
  \emph{Journal of Personality and Social Psychology}, vol.~32, no.~5, p. 922,
  1975.

\bibitem{liebrand1984effect}
W.~B. Liebrand, ``The effect of social motives, communication and group size on
  behaviour in an n-person multi-stage mixed-motive game,'' \emph{European
  Journal of Social Psychology}, vol.~14, no.~3, pp. 239--264, 1984.

\bibitem{messick1968motivational}
D.~M. Messick and C.~G. McClintock, ``Motivational bases of choice in
  experimental games,'' \emph{Journal of Experimental Social Psychology},
  vol.~4, no.~1, pp. 1--25, 1968.

\bibitem{liebrand1986might}
W.~B. Liebrand, R.~W. Jansen, V.~M. Rijken, and C.~J. Suhre, ``Might over
  morality: Social values and the perception of other players in experimental
  games,'' \emph{Journal of Experimental Social Psychology}, vol.~22, no.~3,
  pp. 203--215, 1986.

\bibitem{griesinger1973toward}
D.~W. Griesinger and J.~W. Livingston~Jr, ``Toward a model of interpersonal
  motivation in experimental games,'' \emph{Behavioral science}, vol.~18,
  no.~3, pp. 173--188, 1973.

\bibitem{DanielaRusSVO}
W.~Schwarting, A.~Pierson, J.~Alonso-Mora, S.~Karaman, and D.~Rus, ``Social
  behavior for autonomous vehicles,'' \emph{Proceedings of the National Academy
  of Sciences}, vol. 116, no.~50, pp. 24\,972--24\,978, 2019.

\bibitem{sutton2018reinforcement}
R.~S. Sutton and A.~G. Barto, \emph{Reinforcement learning: An
  introduction}.\hskip 1em plus 0.5em minus 0.4em\relax MIT press, 2018.

\bibitem{highD}
R.~Krajewski, J.~Bock, L.~Kloeker, and L.~Eckstein, ``The highd dataset: A
  drone dataset of naturalistic vehicle trajectories on german highways for
  validation of highly automated driving systems,'' in \emph{International
  Conference on Intelligent Transportation Systems}.\hskip 1em plus 0.5em minus
  0.4em\relax IEEE, 2018, pp. 2118--2125.

\bibitem{TomiPolyLaneChange}
I.~Papadimitriou and M.~Tomizuka, ``Fast lane changing computations using
  polynomials,'' in \emph{Proceedings of the 2003 American Control Conference,
  2003.}, vol.~1.\hskip 1em plus 0.5em minus 0.4em\relax IEEE, 2003, pp.
  48--53.

\end{thebibliography}
\end{document}